\documentclass[a4]{ecai2014}
\usepackage{times}
\usepackage{graphicx}
\usepackage{latexsym}
\usepackage{amsmath}
\usepackage{amssymb}
\usepackage{amsmath}
\usepackage{mathrsfs}

\DeclareGraphicsExtensions{.pdf}

\usepackage{xspace}

\begin{document}

\newtheorem{mytheorem}{Proposition}
\newcommand{\myproof}{\noindent {\bf Proof:\ \ } \\}
\newcommand{\myqed}{\mbox{$\diamond$}}
\newcommand{\mytrue}{\mbox{\it true}}
\newcommand{\myfalse}{\mbox{\it false}}
\newcommand{\mydiv}{\mbox{\rm div}}
\newcommand{\mymod}{\mbox{\scriptsize \rm \ mod \ }}
\newcommand{\mymin}{\mbox{\rm min}}
\newcommand{\mymax}{\mbox{\rm max}}
\newcommand{\calFR}{{\cal FR}}
\newcommand{\calF}{{\cal F}}
\newcommand{\calC}{{\cal C}}
\newcommand{\calA}{{\cal A}}
\newcommand{\calL}{{\cal L}}
\newcommand{\dom}{{dom}}
\newcommand{\real}{\ensuremath{\mathbb{R}}}
\newcommand{\nat}{\ensuremath{\mathbb{N}}}
\newcommand{\ZZ}{\ensuremath{\mathbb{Z}}}
\newcommand{\upw}{UPW}
\newcommand{\unw}{UNW}
\newcommand{\wpw}{WPW}
\newcommand{\wnw}{WNW}

\newcommand{\mf}{{\mathscr F}}
\newcommand{\mc}{\mathcal C}
\newcommand{\ra}{\rightarrow}

\newcommand{\set}{\mathcal}
\newcommand{\myset}[1]{\ensuremath{\mathcal #1}}
\newcommand{\myOmit}[1]{}
\newcommand{\tighter}{\mbox{$\preceq$}}
\newcommand{\stighter}{\mbox{$\prec$}}
\newcommand{\incomparable}{\mbox{$\bowtie$}}
\newcommand{\equivalent}{\mbox{$\equiv$}}

\newcommand{\reg}{\mbox{$RE$}}
\newcommand{\mreg}{\mbox{$MR$}}
\newcommand{\hprs}{\mbox{$HPRS$}}
\newcommand{\reglo}{\mbox{$LO$}}
\newcommand{\ps}{\mbox{$PS$}}
\newcommand{\tc}{\mbox{$ST$}}
\newcommand{\among}{\mbox{$AD$}}
\newcommand{\lse}{\mbox{$LG$}}
\newcommand{\lser}{\mbox{$LG_R$}}
\newcommand{\cs}{\mbox{$CS$}}
\newcommand{\csdc}{\mbox{$CS_{DC}$}}
\newcommand{\fl}{\mbox{\sc $FB$}}
\newcommand{\flS}{\mbox{\sc $FB_S$}}
\newcommand{\amongS}{\mbox{$AD_S$}}

\newcommand{\gsc}{\mbox{\sc Gsc}}
\newcommand{\gcc}{\mbox{\sc Gcc}}
\newcommand{\GCC}{\mbox{\sc Gcc}}
\newcommand{\AllDifferent}{\mbox{\sc AllDifferent}}

\newcommand{\nina}[1]{{#1}}

\newcommand{\SLIDE}{\mbox{\sc Slide}}
\newcommand{\SLIDINGSUM}{\mbox{\sc SlidingSum}}
\newcommand{\REGULAR}{\mbox{\sc Regular}}
\newcommand{\TABLE}{\mbox{\sc Table}}
\newcommand{\CIRCREGULAR}{\mbox{$\mbox{\sc Regular}_{\odot}$}}
\newcommand{\WRAPREGULAR}{\mbox{$\mbox{\sc Regular}^{*}$}}
\newcommand{\STRETCH}{\mbox{\sc Stretch}}
\newcommand{\INCSEQ}{\mbox{\sc IncreasingSeq}}
\newcommand{\INC}{\mbox{\sc Increasing}}
\newcommand{\lseX}{\mbox{\sc Lex}}
\newcommand{\NFA}{\mbox{\sc NFA}}
\newcommand{\DFA}{\mbox{\sc DFA}}

\newcommand{\PRECEDENCE}{\mbox{\sc Precedence}}
\newcommand{\lseXVAR}{\mbox{\sc LexLeader}}
\newcommand{\lseXGENSET}{\mbox{\sc SetGenLexLeader}}
\newcommand{\lseXSETVAR}{\mbox{\sc SetLexLeader}}
\newcommand{\lseXMSETVAR}{\mbox{\sc MSetLexLeader}}
\newcommand{\lseXSETVAL}{\mbox{\sc SetValLexLeader}}
\newcommand{\lseXVAL}{\mbox{\sc ValLexLeader}}
\newcommand{\lseXVALVAR}{\mbox{\sc GenLexLeader}}
\newcommand{\VALVARLEX}{\mbox{\sc ValVarLexLeader}}
\newcommand{\NVALUES}{\mbox{\sc NValues}}
\newcommand{\USES}{\mbox{\sc Uses}}
\newcommand{\COMMONG}{\mbox{\sc Common}}
\newcommand{\CARDPATH}{\mbox{\sc CardPath}}
\newcommand{\RANGE}{\mbox{\sc Range}}
\newcommand{\ROOTS}{\mbox{\sc Roots}}
\newcommand{\AMONG}{\mbox{\sc Among}}
\newcommand{\ATMOST}{\mbox{\sc AtMost}}
\newcommand{\ATLEAST}{\mbox{\sc AtLeast}}
\newcommand{\ATMOSTSEQ}{\mbox{\sc AtMostSeq}}
\newcommand{\ATLEASTSEQ}{\mbox{\sc AtLeastSeq}}
\newcommand{\AMONGSEQ}{\mbox{\sc AmongSeq}}
\newcommand{\SEQUENCE}{\mbox{\sc Sequence}}
\newcommand{\GENSEQUENCE}{\mbox{\sc Gen-Sequence}}
\newcommand{\SEQ}{\mbox{\sc Seq}}
\newcommand{\myelement}{\mbox{\sc Element}}
\newcommand{\LEX} {\mbox{\sc Lex}}
\newcommand{\REPEAT} {\mbox{\sc Repeat}}
\newcommand{\REPEATONE} {\mbox{\sc RepeatOne}}
\newcommand{\STRETCHREPEAT} {\mbox{\sc StretchRepeat}}
\newcommand{\STRETCHONEREPEAT} {\mbox{\sc StretchOneRepeat}}
\newcommand{\STRETCHONEREPEATONE} {\mbox{\sc StretchOneRepeatOne}}
\newcommand{\SETSIGLEX} {\mbox{\sc SetSigLex}}
\newcommand{\SETPREC} {\mbox{\sc SetPrecedence}}

\newcommand{\SOFTATMOSTSEC} {\mbox{\sc SoftAtMostSequence}}
\newcommand{\ATMOSTSEC} {\mbox{\sc AtMostSequence}}
\newcommand{\SOFTATMOST} {\mbox{\sc SoftAtMost}}
\newcommand{\SOFTSEQ} {\mbox{\sc SoftSequence}}
\newcommand{\SOFTAMONG} {\mbox{\sc SoftAmong}}
\newcommand{\SEQCYC}{\mbox{$\mbox{\sc CyclicSequence}$}}

\newcommand{\ATMOSTSEQCYC}{\mbox{$\mbox{\sc AtMostSeq}_{\odot}$}}
\newcommand{\ignore}[1]{}

\newcommand{\ms}{\mathcal S}
\newcommand{\ma}{\mathcal A}
\newcommand{\mv}{\mathcal V}
\newcommand{\rev}{\text{rev}}
\newcommand{\others}{\text{\it Others}}

\newcommand{\vote}[3]{\mbox{$#1 \! \succ \! #2 \! \succ \! #3$}\xspace}
\newcommand{\vvote}[4]{\mbox{$#1 \! \succ \! #2 \! \succ \! #3 \! \succ \! #4$}\xspace}

\newcommand{\new}[1]{{#1}}
\newcommand{\myvec}[1]{\vec{#1}}

\renewcommand{\restriction}{\mathord{\upharpoonright}}
\newcommand{\votingRule}{\textsc{Maj}2\xspace}

\pagestyle{plain}

\title{The {PeerRank} Method for Peer Assessment}

\author{
Toby Walsh
\institute{NICTA and UNSW, Sydney, Australia.
{NICTA is funded by
the Australian Government as represented by
the Department of Broadband, Communications and the Digital Economy and
the Australian Research Council. The author is also supported
by AOARD Grant FA2386-12-1-4056.}}}


\maketitle

\begin{abstract}
We propose the PeerRank method for peer assessment.
This constructs a grade for an agent
based on the grades proposed by the agents evaluating the agent. 
Since the grade of an agent is a measure of their ability 
to grade correctly, the PeerRank method weights 
grades by the grades of the grading agent. 
The PeerRank method also provides an incentive
for agents to grade correctly.
As the grades of an agent
depend on the grades of the grading agents,
and as these grades themselves depend on the
grades of other agents, we define the PeerRank method
by a fixed point equation similar
to the PageRank method for ranking web-pages. 
We identify some formal properties of the PeerRank method
(for example, it satisfies axioms of unanimity,
no dummy, no discrimination and symmetry),
discuss some examples, compare with 
related work and evaluate the performance
on some synthetic data. 
Our results show considerable promise, reducing the error in grade predictions by a factor of 2 or more in many cases over the natural baseline of averaging peer grades.
\end{abstract}




\section{INTRODUCTION}

We consider how to 
combine together peer assessments of
some work to construct an overall evaluation of
this work. 
An important application of our proposed
framework is to evaluation in massive
open online courses (MOOCs). In such a setting,
it may be impractical to offer anything
but automated marking (where this 
is possible) or peer assessment (e.g. for 
essays where this might not be possible). 
%
Another application of the proposed framework
is to peer assessment of grant applications. 
Often there is only a small pool of experts
who are capable of reviewing grant
applications in a particular sub-area. In many cases, these people
have also submitted grant applications themselves.
It is natural therefore to consider designing a mechanism
in which those people submitting proposals also
review them. 

Unfortunately, peer assessment suffers from several fundamental 
problems. First, how can we provide an incentive
to agents to assess their peers well? 
Second, as peers may have different
expertise, how do we compensate for any
unintentional biases that peer assessment may introduce? 
Third, as peers may not be 
disinterested in the
outcome, how do we compensate for any
intentional biases that peer assessment may introduce? 
In this paper, we view this as a mechanism design
problem in which we look to provide
incentives for peers to assess well,
as well as a means to try to compensate for
any biases. 


Our main contribution
is to propose the PeerRank method for peer assessment.
This constructs a grade for an agent
based on the grades proposed by the agents evaluating the agent. 
The PeerRank method makes two basic 
assumptions about how peer grades should be
combined.
First, it supposes that the grade of an agent is a measure of their 
ability to grade correctly. Hence,
grades are weighted by the grades of
grading agents. 
Second, agents should be rewarded for grading 
correctly. This gives agents an incentive
to provide accurate peer assessments. 
%
%
We identify some formal properties of the PeerRank method.
We also evaluate the performance
on some synthetic data. As our method
favours consensus, it is most suited
to domains where there are objective
answers but the number of agents is too large
for anything but peer grading. 

We hope that this work will
encourage others to consider peer assessment
from a similar (social choice) perspective. There are other
axiomatic properties we could formalise and study.
For instance, the PeerRank rule is not
monotonic. Increasing the grade for an agent
can hurt an agent if they thereby
receive a bigger proportion of their support from 
agents that grade poorly. On the other
hand, the PeerRank rule likely satisfies a more complex form 
of monotonicity, in which reducing the error
in the grade of an agent only ever helps that 
agent. We expect there are important 
axiomatic results to be obtained about peer
assessment.
Finally, an interesting extension would
be to return a distribution or interval of 
grades, reflecting the uncertainty in the 
estimate. 
  

\section{PEER RANK RULE}

We suppose there are $m$ agents, and agent $j$ provides
a grade $A_{i,j}$ for the exam of agent $i$. 
Grades are normalised into the interval $[0,1]$. 
We suppose agents grade their own work but 
this can be relaxed. In addition, as we show in the
experimental section, the proposed PeerRank rule
is relatively insensitive to any bias that
an agent might have towards grading their own work or
that of other agents. 
The grade of each agent
is constructed from the grades of 
the agents evaluating the agent. 
Since the  grade is a measure of their
ability to grade correctly, 
we weight the grade an agent gives another agent
by their own grade. The grade of an agent
is thus the weighted average of the grades
of the agents evaluating the agent. 
Now the grades
of the agents evaluating an agent
are themselves the weighted averages of the grades
of the agents evaluating the agents. Hence we
set up a set of equations and look
for the fixed point of the system. 
This is reminiscent of
the problem faced by the PageRank algorithm \cite{pagerank}.
In PageRank, web-pages
are ranked according to the ranks of the web-pages that link
to them, these ranks depend
on the ranks of the web-pages that
like to them, and so on. 

Let $X^n_{i}$ be the
grade of agent $i$ in the $n$th iteration
of the PeerRank rule and $0 < \alpha < 1$.
We define the grades at each iteration as follows:
\begin{eqnarray*}
X^0_i & = & \frac{1}{m} \sum_j A_{i,j} \\
X^{n+1}_i & = & (1-\alpha) . X^n_i + \frac{\alpha}{\sum_j X^n_j} \sum_j  X^n_j . A_{i,j}
\end{eqnarray*}
The last term is merely the average grade of an agent
weighted by the current grades. 
The PeerRank grades are the fixed point 
of these set of equations. 
Note that whilst we start with the (unweighted) average
grade, this choice is not very critical and 
we will typically reach the same fixed point with other 
initial seeds. Similarly, the choice of the 
exact value of $\alpha$ is not critical
and largely affects the speed of convergence. 
This is because the fixed point is
an eigenvector of the grade matrix $A$. 

\begin{mytheorem}[Fixed point]
The PeerRank rule returns grades that are
an eigenvector of the grade matrix $A$. 
\end{mytheorem}
\myproof
In matrix notation, at the fixed point,
we have:
$$X = (1-\alpha)X + \frac{\alpha}{|X|} A.X$$
That is, 
$$X = X - \alpha X + \frac{\alpha}{|X|} A.X$$
Rearranging and cancelling  gives:
$$\frac{\alpha}{|X|} A.X = \alpha X$$
Dividing by $\alpha$ and letting $\lambda = |X|$,
we get:
$$A.X = \lambda X$$
\myqed

\myOmit{
We first discuss how the PeerRank rule works
on some simple grade matrices like the identity
matrix, before
considering its (axiomatic) properties.}

\section{SOME EXAMPLES}

To illustrate how the PeerRank rule works on some
simple cases, we consider a few examples. 

\subsection*{Unanimous grade matrix}

Suppose that every entry in the grade matrix $A$ 
is the grade $k$ with $0 \leq k \leq 1$. Now an
eigenvector of $A$, and the PeerRank
solution assigns each agent with this 
grade $k$. The weighted average of identical grades
is always the same whatever the weights. 
This is what we might expect.
The grade matrix tells us nothing more than this.

\subsection*{Identity grade matrix}

Suppose the grade matrix $A$ is the identity
matrix. That is, each agent gives themselves
the maximum grade 1, and every other agent
the minimum grade 0. Now an
eigenvector of $A$, and the PeerRank
solution assigns each agent with the 
average grade $\frac{1}{m}$. 
Again, this is what we might expect.
The grade matrix tells us nothing more than 
all agents are symmetric, and so dividing
the mark between them might seem reasonable. 

\subsection*{Bivalent grade matrices}

Suppose that agents partition into two
types: good and bad. The good agents 
give a grade of 1 to other good agents,
and 0 to bad agents. The bad agents 
gives a grade of 1 to every agent. 
In each iteration of the
PeerRank method, the grades of the good agents remain
unchanged at 1. 
On the other hand, the grades of the bad agents 
monotonically
decrease towards their fixed point at 0. 
We also get the same fixed point if the 
bad agents give a grade of 0 to every
agent besides themselves (irrespective
of the grade that they give themselves).
Again, this is what we might expect.
The PeerRank method identifies the good
and bad agents, and rewards them appropriately. 

\section{PROPERTIES}

The PeerRank rule has a number of
desirable (axiomatic) properties. 
Several of these properties (e.g. no dummy
and no discrimination) are properties
that have been studied by 
in peer selection of a prize 
\cite{hmeconometrica2013}. 
First, we argue that the PeerRank rule returns
a normalised grade. 

\begin{mytheorem}[Domain]
The PeerRank rule returns grades in $[0,1]$. 
\end{mytheorem}
\myproof
Clearly $X^n_i \geq 0$ for all $n$ as it is the
sum of two terms which are never negative.
We prove that $X^n_i \leq 1$ by induction on $n$ 
In the base case, $X^0_i \leq 1$ as it
is the average of terms which are themselves
less than or equal to 1. 
In the step case, suppose $0 \leq X^n_i \leq 1$ for all $i$.
Let $X^n_i = 1 - \epsilon$ where $0 \leq \epsilon \leq 1$.
Then 
\begin{eqnarray*}
X^{n+1}_i & = & (1-\alpha)(1-\epsilon) + \frac{\alpha}{\sum_j X^n_j}
\sum_j X^n_j . A_{i,j}  \\
& \leq & 1 -\alpha - \epsilon (1-\alpha) +
\frac{\alpha}{\sum_j X^n_j} \sum_j X^n_j  \\
& = & 1 -\alpha - \epsilon (1-\alpha) + \alpha \\
& = & 1 - \epsilon (1 - \alpha) \ \leq 1
\end{eqnarray*} 
Note that these bounds are reachable. 
In particular, if all peer grades are 0 (1) then the
PeerRank rule gives every agent this grade. 
\myqed

\noindent
Next we argue that 
if all agents give
an agent the same grade then this is 
their final grade. 

\begin{mytheorem}[Unanimity]
If all agents give an agent the grade
$k$ then the PeerRank rule gives
this grade $k$ to the agent. 
\end{mytheorem}
\myproof
Suppose all agents give agent $i$ the grade $k$. 
Consider the $i$th component of the fix 
point equation:
$$X_i = (1-\alpha).X_i + \frac{\alpha}{\sum_j X_j} \sum_j X_j . A_{i,j} $$
Rearranging gives:
$$\alpha X_i = \frac{\alpha}{\sum_j X_j} \sum_j X_j . A_{i,j} $$
Dividing by $\alpha$ and multiplying up the fraction gives:
\begin{eqnarray*}
(\sum_j X_j) . X_i & = & \sum_j X_j . A_{i,j} \\
& = & (\sum_j X_j) . k
\end{eqnarray*}
Dividing by the common term, $\sum_j X_j$, we
get:
$X_i = k$.
\myqed

\noindent 
The PeerRank rules also satisfies a no discrimination axiom.
Every vector of grades is possible. 

\begin{mytheorem}[No discrimination]
Given any vector of grades, there exists
a grade matrix with which the PeerRank rule 
returns this vector. 
\end{mytheorem}
\myproof
Suppose we want agent $i$ to get the grade $k_i$. 
Then we construct the grade matrix with $A_{i,j} = k_i$
and appeal to unanimity.
\myqed

\noindent 
The PeerRank rules also satisfies a no dummy axiom
since every agent has some influence over the final
grade. 

\begin{mytheorem}[No dummy]
There exist two grade matrices which
differ in just the grades assigned
by one agent for which PeerRank returns
different final grades. 
\end{mytheorem}
\myproof
Consider the grade matrix in which
every agent gives the maximum grade of 1 to every
other agent, and the grade matrix
which is identical except agent $i$
gives every agent the minimum grade of 0. 
Then PeerRank gives a grade of
1 to agent $i$ in the first case and 0 in the second. 
Hence $i$ is not a dummy. 
\myqed

\noindent 
The PeerRank rules also satisfies a simple 
symmetry axiom. 

\begin{mytheorem}[Symmetry]
If we swap the grades of two agents and the
grades that the two agents are given then
the PeerRank rule swaps the grades assigned
to the two agents. 
\end{mytheorem}


\noindent
It is also interesting to identify properties
that the PeerRank rule does not have. For example,
it is not impartial. Your grades of others do affect 
your own final grade. As a second example, it is not
anonymous. It does matter who gives you a grade. 
It is better to get a good grade from an agent
who themself receives good grades than from
an agent who themself receives poor grades. 

\section{GENERALIZED PEERRANK}

The PeerRank rule proposed so far does not 
incentivize agents to evaluate other agents or even
themselves accurately. 
We therefore add an additional term to
provide such an incentive. Suppose $\alpha$ and $\beta$ are
parameters with $\alpha + \beta \leq 1$. Then
we define the generalised PeerRank rule recursively
by the following equation:
\begin{eqnarray*}
X^{n+1}_i & = & (1-\alpha-\beta) . X^n_i + \frac{\alpha}{\sum_j X^n_j} . \sum_j X^n_j . A_{i,j} + \\
 &  &  \ \ \frac{\beta}{m} . \sum_j 1-|A_{j,i}-X^n_j| 
\end{eqnarray*}
This degenerates
to the earlier form of the rule
when $\beta=0$. 
The new term measures the normalised absolute error in the
grades given by an agent. This is similar
to the reward given in the recent mechanism
for reviewing NSF proposals in the SSS program
\cite{nsf1}. 
The agent ``receives'' a credit towards
their grade of $\beta$ times this normalised error.

If $A_{j,i} = X^n_j$ for all $j$ then the grades assigned
by an agent
are exact and we add $\beta$ to their score.
If $|A_{j,i} - X^n_j|=1$ for all $j$ then the grades
assigned by an agent
are completely wrong (either the agent gives a grade of 1
when it should be 0 or vice versa). In this case,
their grade is reduced by a factor $\beta$ for evaluating
incorrectly. 

The generalised PeerRank rule continues to
satisfy the domain, no discrimination,
no dummy, and symmetry properties. 
For no discrimination, and no dummy, 
we can just set $\beta=0$ and appeal
to the previous results. For the domain
property, we need to prove afresh that
the additional term 
cannot take us outside the interval $[0,1]$. 

\begin{mytheorem}[Domain]
The generalized PeerRank rule returns grades in $[0,1]$. 
\end{mytheorem}
\myproof
Clearly $X^n_i \geq 0$ for all $n$ as it is the
sum of terms which are not negative. 
We prove that $X^n_i \leq 1$ by induction on $n$ 
In the step case, suppose $0 \leq X^n_i \leq 1$ for all $i$.
Let $X^n_i = 1 - \epsilon$ where $0 \leq \epsilon \leq 1$.
Then 
\begin{eqnarray*}
X^{n+1}_i & = & (1-\alpha-\beta)(1-\epsilon) + \frac{\alpha}{\sum_j X^n_j}
\sum_j X^n_j . A_{i,j} + \\
& & \ \ \ \ \frac{\beta}{m}. \sum_j 1 - |A_{j,i} -X^n_j| \\
& \leq & 1 -\alpha - \beta - \epsilon (1-\alpha - \beta) +
\frac{\alpha}{\sum_j X^n_j} \sum_j X^n_j  +  \\
& & \ \ \ \ \ \frac{\beta}{m} \sum_j 1 \\
& \leq & 1 -\alpha - \beta - \epsilon (1-\alpha - \beta) + \alpha + \beta \\
& \leq & 1 - \epsilon (1 - (\alpha + \beta))
\end{eqnarray*}
Recall that $\alpha + \beta \leq 1$ and $\epsilon \geq 0$. 
Thus, $\epsilon (1 - (\alpha + \beta)) \geq 0$.
Hence $X^{n+1}_i \leq 1$. 
\myqed

To demonstrate the impact of the
new term that encourages accurate
peer grading, we consider again the simple grade matrices
considered previously. 

\subsection*{Unanimous grade matrix}

Suppose every entry in the grade matrix $A$ is
the grade $k$. Now 
the generalised PeerRank
solution assigns each agent with a
grade greater than or equal to $k$
(with equality when $m=1$, $k=1$ or $\beta=0$).
Grades increase above $k$
as agents receive some credit for grading
accurately. 

\subsection*{Identity grade matrix}

Suppose the grade matrix $A$ is the identity
matrix. That is, each agent gives themselves
the maximum grade 1, and every other agent
the minimum grade 0. Now the generalised PeerRank
solution assigns each agent a
grade greater than or equal to $\frac{1}{m}$
(with equality when $m=1$ or $\beta=0$). Grades 
are larger than $\frac{1}{m}$ 
as agents receive credit for grading
themselves semi-accurately. 

\subsection*{Bivalent grade matrices}

Suppose that agents partition into two
types: good and bad. The good agents 
give a grade of 1 to the good agents,
and 0 to the bad agents. The bad agents 
gives a grade of 1 to every agent. 
Now the generalised PeerRank method
give the good agents a grade less than
or equal to 1, and the bad agents 
a grade more than
or equal to 0. The bad agents get 
some credit for grading the good agents
(semi-)accurately. This means that
the grade of the bad agents by the good agents
was a little too harsh, and their
own grade suffers. 

\section{EXPERIMENTAL EVALUATION}

We tested the performance of the generalised
PeerRank rule on some synthetic data. 
In all experiments, we set $\alpha=\beta=0.1$.
Results are, however, relatively insensitive to the actual
choice of $\alpha$ or $\beta$. 
Based on the promise shown in these experiments, we are
currently preparing a real world test with 
undergraduate students. Our typical 
experimental setup is 10 agents who 
give an integer mark to each other
of between 0 and 10, and an actual
mark of between 0 and 100. 
Therefore a simple baseline against which
we compare is the sum of these peer graded marks
(or equivalently the average of the
normalised peer grade). We denote this 
as the AVERAGE rule. 

We studied a number of different 
distributions of marks amongst 
the agents (e.g. binomial,
normal, uniform). These are discussed in more
detail in the next section. 
We also need a marking model to determine
how well the grading agents
grade. We used a simple model based on each mark
being awarded independently with a probability
given by the grade of the grading agent. In our experiments, this 
means that the agents are effectively answering 10 questions,
that the probability of each of these
questions being answered correctly is
their actual grade, and that the 
probability of each of these questions being
graded correctly is the grade of the 
grading agent. This gives a distribution
of marks that is the sum of two binomials. 

For instance, if the actual mark of an agent is 
62 out of 100, then we expect
their peer grade to be (on average)
6 out of 10. Suppose their work is marked
by an agent whose actual mark is 
72 out of 100. On the 6 questions that the agent
is expected to get right, we suppose that each is marked
correctly by this peer with probability 0.72.
This gives a binomial distribution
of 6 marks with a probability of 0.72.
On the 4 questions that they got
wrong, we suppose also that they are marked
correctly by this peer (as false) with probability 0.72, and
incorrectly (as true) with probability 1-0.72.
This gives again a binomial distribution
of 4 marks with a probability of 1-0.72.
Hence, the final mark given to the agent by their peer is the sum
of these two binomial distributions:
$bin(6,0.72)+bin(4,1-0.72)$ where $bin(m,p)$ is
a binomial distribution of $m$ trials
with probability $p$. 

We tried other marking models 
including normally distributed peer grades with
a standard deviation that is inversely
proportional to the grade of the marking 
agent, and uniform distributed peer grades
with a range that is also
inversely
proportional to the grade of the marking 
agent. As we obtained similar results 
with these other marking models, we focus
here on our simple sum of binomials model.

\subsection{Mark distributions}

We begin with a simple binomial distribution
of marks. We let the actual mark of the agents
be a binomial distribution of 100 trials with a given
probability $p$. 
In Figure 1, we plot
the RMSE (root mean square error) of the predicted
mark as a percentage of the 100 marks
for varying $p$. Thus a RMSE of 5\% means that the PeerRank
grade is off with a root mean square
error of 5 marks (out of the 100
possible marks). If we map back onto grades out of 
10, this means we are off by less than half a grade.
For $p>0.6$ (in other
words, for where the marks are
typically above 60 out of 100), the generalised PeerRank
method outperforms simply averaging the peer
grades. For $p>0.65$, the error is
4\% or less. This compares well
with the error returned by
simply averaging the peer grades
(which is mostly above 10\% 
in this region). 
Note that for PeerRank to get any useful signal out of the 
data, we need $p>0.5$. 
At $p=0.5$, we will often 
answer (or mark) an exam 
just as well by tossing a coin. 
With the PeerRank method,
we need the exam to be informative (that is,
to have $p > 0.6$), to be able to extract
much information from the grade matrix.

\begin{figure}[htbp]
\vspace{-0.5em}
\begin{center}
\includegraphics[scale=0.9]{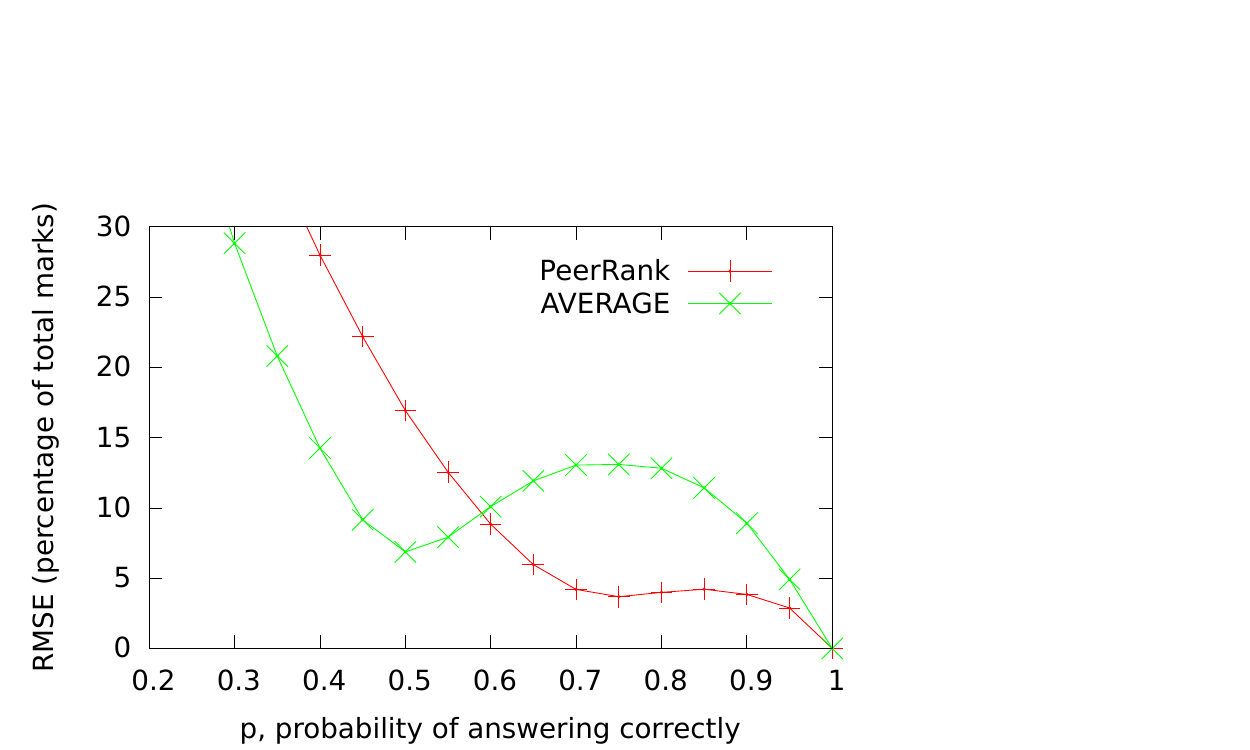}
\end{center}
\caption{Performance of the generalised PeerRank
method on marks coming from a binomial distribution
with parameter $p$.}
\vspace{-0.5em}
\end{figure}

We next turned to a normal distribution of marks.
This permits us to study the impact of 
increasing the standard deviation in marks.
With the previous binomial distribution of marks, 
the standard deviation is $\sqrt{100p(1-p)}$
which is fixed by $p$. 
In Figure 2, we plot
the error in the predicted mark 
for varying standard deviations. 
The mean grade is fixed at 70 marks
out of 100. 
We again see that the generalised PeerRank
method outperforms simply averaging
peer grades except when there is a 
very large standard deviation in 
marks. 

\begin{figure}[htbp]
\vspace{-0.5em}
\begin{center}
\includegraphics[scale=0.9]{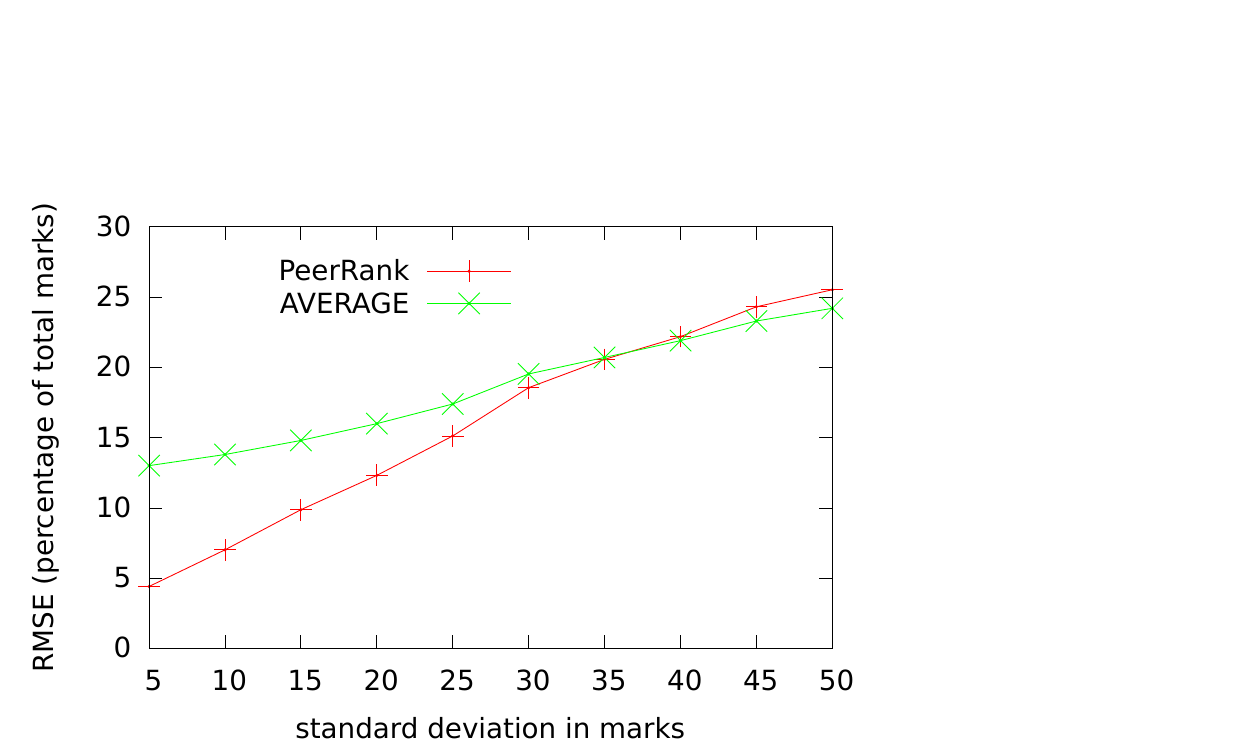}
\end{center}
\caption{Performance of the generalised PeerRank
method on marks from a normal distribution
with a mean of 70 as we vary the standard
deviation.}
\vspace{-2em}
\end{figure}

Finally, we consider a simple uniform
distribution of marks. We suppose 
that every mark from $lo$ to 100 is
equally likely. 
\begin{figure}[htbp]
\vspace{-0.5em}
\begin{center}
\includegraphics[scale=0.9]{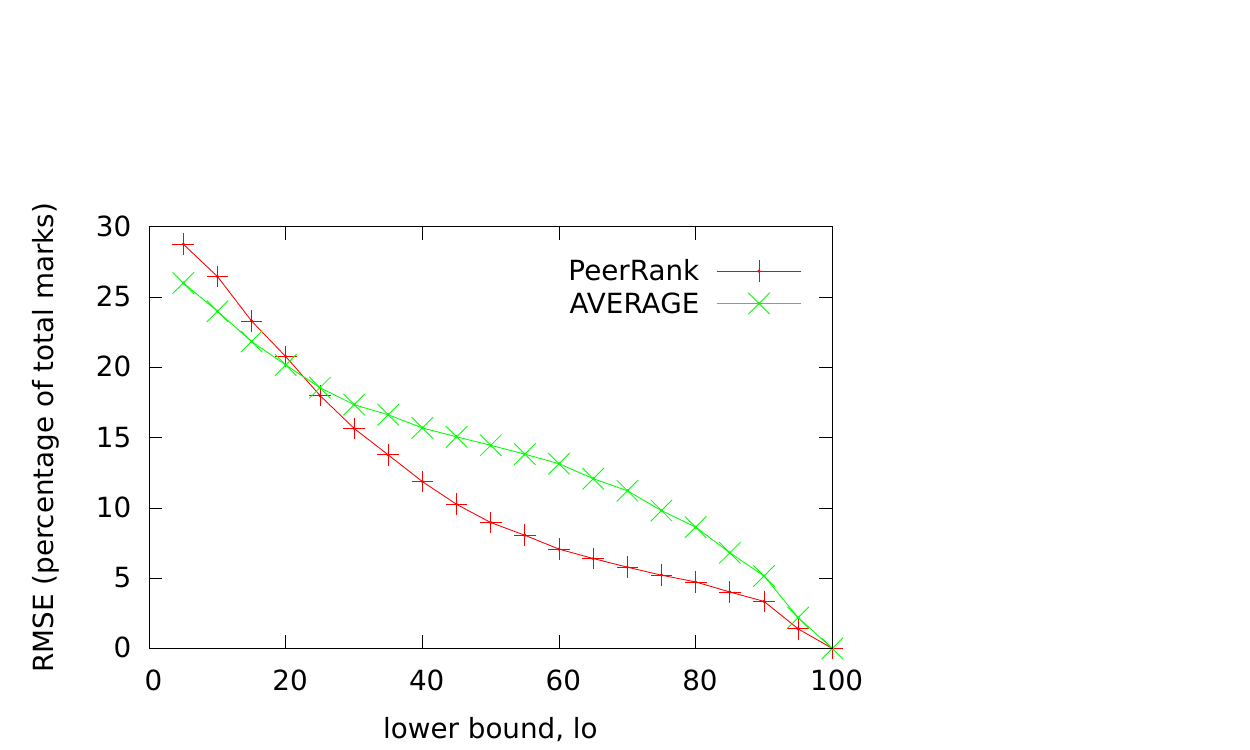}
\end{center}
\caption{Performance of the generalised PeerRank
method on marks coming from the uniform distribution
$[lo,100]$ for varying lower bound $lo$. }
\vspace{-0.5em}
\end{figure}
In Figure 3, we plot
the error in the predicted grade 
whilst
we vary $lo$, the lowest possible mark. 
From $lo>20$, the generalised PeerRank
method outperforms simply averaging the 
peer grades. For $lo \geq 50$, the error 
is less than 10\%.
As with binomially distributed marks, 
the exam needs to be informative (that is,
for marks to be above 50), to be able to extract
information from the grade matrix.

\subsection{Group size}

So far, we have supposed that there are 10 agents
who grade each other. We next consider the
impact that the size of this group has on the 
accuracy of the peer grades. 
We therefore ran an experiment in which we
varied the number of agents peer marking. 
We again use a binomial distribution of marks
with a mean of 70. 
With 5 or more agents,  the error of
the generalised PeerRank method was less than 5\%
and was half or less of that of simply averaging
the peer grades. With 10 to 20 agents, 
the error of the generalised PeerRank method was 
less than a third of that of simply averaging
the peer grades. These results suggest
that the PeerRank method does not need many peer grades in order
to obtain an accurate result. Ideally, we 
need around 10 grades for each agent, but
even with just 5 grades, we are often able to 
obtain acceptable results.


\subsection{Biased marks}

Peer grades may be systematically biased.
For instance, students may collude and 
agree to grade each other generously. 
Even if there is no explicit collusion,
there are studies which suggest that students
grade each other generously 
(e.g. \cite{kpdjeb95}). 
To study this, we inflate or deflate the mean
of the peer grades by a factor $r$. 
For instance, if $r=1.1$ then the
mean peer mark is increased by 10\%. 
On the other hand, if $r=0.9$ then the
mean peer mark is decreased by 10\%. 
We again use a binomial distribution of 
actual marks with a mean of 70. 

\begin{figure}[htbp]
\vspace{-0.5em}
\begin{center}
\includegraphics[scale=0.9]{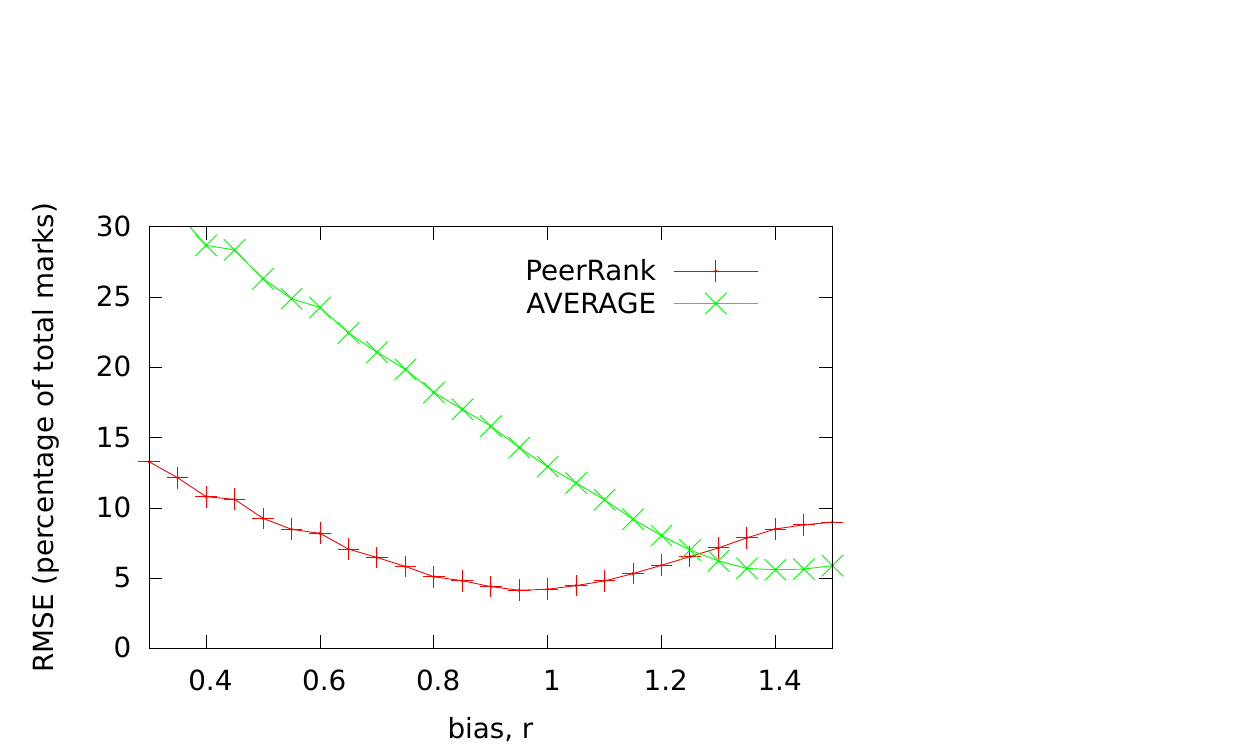}
\end{center}
\caption{Performance of the generalised PeerRank
method on marks coming from a binomial
distribution with a mean of 70 (i.e. $p=0.7$) as we vary the bias
in the peer marking. For $r>1$, peers 
return inflated marks. For $r<1$, peers
return deflated marks. }
\vspace{-0.5em}
\end{figure}

In Figure 4, we plot
the RMSE of the predicted grade again 
as a percentage of the 100 marks whilst
we vary the bias in peer grades. 
For $0.75 \leq r \leq 1.25$, 
the error of the generalised PeerRank method is 
5\% or less of the total marks. That is,
we are able to tolerate a bias of 
25\% in peer grades without
significantly increasing the error. 
These results suggest that
the generalised PeerRank method has some
robustness against bias. 
The minimum in errors for averaging peer grades 
at around $r=1.5$ is likely 
an artefact of the model. Averaging peer grades tend to
under-estimate the actual grade.
Therefore a positive bias on the peer grades
tends to reduce this. 

\section{RELATED WORK}

There is a large literature on peer assessment
but the focus is mostly on pedagogical aspects
of peer assessment (for example, how peer
assessment itself contributes to the learning
experience). There is less literature on how best
to combine peer assessments together. 
Many of the peer assessments systems being used in
practice today often have simple
and rather ad hoc mechanisms for combining
together peer assessments. 
In addition to multiple choice questions and
computer grading, 
peer assessment has been used on a number of Coursera
courses. Students first train
with a grading rubric. To get feedback on their
own work, a student has to grade five essays.
The student then receives peer grades from five
other students. Using machine learning algorithms, Piech {\it et al.}
have estimated and corrected for grader biases and reliabilities
\cite{edm2013}. 
They demonstrate
significant improvement in grading accuracy on real
world data of over 63,000 peer grades. 
Their models are probabilistic so give 
a belief distribution over grades
(as opposed to a single score) for each student.
\myOmit{
It would be very interesting to compare their
performance with that of the methods proposed here. 
One difference between their methods
and those proposed here is that
only their third and most complex model links
the final grade of a student with 
the quality of their peer assessments. 
By comparison, this link is very direct
in our model. }

PEAS is a new peer evaluation extension for the EdX open source
MOOC platform \cite{peas}. 
Students are incentivized to grade accurately
by a calibration method that constructs
an incentive score based 
on the accuracy of their grading.
To improve review quality, 
students are divided into groups based on
this incentive score, and each assignment
is peer graded by one student from each group. 
A simple normalisation of grades is also performed 
to reduce bias in peer grading. Expert
grading can be used to resolve
discrepancies in peer grades, and to
provide training data for Machine Learning
algorithms to grade automatically. 
An important difference with our
work is that the calibration in PEAS is 
just once, whilst PeerRank potentially uses multiple
rounds of adjustment of grades. 

\myOmit{
Calibrated Peer Review is a 
web-based application for peer reviewing 
from UCLA \cite{cpr}.
After completing an essay, 
students calibrate their marks by grading 
three essays provided by the instructor 
using a multiple-choice rubric.
Based on this calibration, students
are assigned a Reviewer Competency Index (RCI) which 
is used as a weighted multiplier for the 
grades that they assign to other students. 
Each student then grades three essays of their peers
with the rubric. The peer review task is scored
by how well the rating by an individual reviewer of the essay matches 
the weighted rating of the
essay. Finally, students complete a self-evaluation of 
their own essay which is scored by how
well they match the weighted review scores of their peers.
The final grade returned to students is
just a weighted sum of their peer grade, 
and grades based on their calibration accuracy, and their
peer and self grading accuracy. 
As with PEAS, an important difference with our
work is that the calibration in Calibrated Peer Review is 
just once. 

Peer assessment has also been used to mark
group projects. For instance, Conway {\it et al.}
report their experience in marking group projects
where other students in the class decide on
a mark for the project, and then the
group members themselves decide on how these marks are
divided using
ratings by each other of their individual
contributions \cite{ckswaehe93}. This division
is based on a complex two part formula
which takes into account both
the frequency with which an individual
was mentioned to have contributed the most and
a zero to four scale of effort. 
Again a significant difference with
this work and ours is that there is no
link between the final grade of a student
and the quality of their peer assessments. 
}

One of the closest works to ours is a 
peer reviewing mechanism 
being piloted by the National Science Foundation (NSF)
for the Signal and Sensing Systems (SSS) program.
The mechanism is designed to help deal
with an increase in proposals which is putting
an increasing stress on the grant reviewing process.
This increase in proposals has led to
a degradation in the quality of reviews,
as well as a shrinking pool of qualified
but non-conflicted reviewers. 
The NSF has therefore decided to pilot a 
mechanism 
for peer review that is adapted from one first
proposed by Merrifield and Saari \cite{nsf1}.
\myOmit{
This mechanism works as follows. Each subfield
of the SSS program attracts between 25 and 40 proposals. 
Each applicant is required to rank 7 other
proposals, ignoring those on which there is 
some conflict of interest. This number was chosen
to be small enough not to be a burden, but 
large enough to discourage submission of multiple
low quality proposals, and to give a reliable
final ranking. Each ranking is converted into a 
Borda score (so that the $i$th ranked proposal
gets a score of $7-i$) and these scores are
summed. This sum is then used to rank all
proposals. 
}
To incentivize applicants 
to review well, and to deter strategic
ranking, reviewers receive additional
score for reviewing well which can 
increase them a maximum of 2 places in the
final ranked list. 
%
This mechanism is somewhat
different to ours as the NSF mechanism ranks
proposals, whilst our
mechanism returns a grade. 
In our mechanism, 
final grades 
returned 
may not totally rank 
the proposals. 


Another work which is close to ours is the
CrowdGrader mechanism for peer evaluation
from UC Santa Cruz \cite{crowdgrader}.
CrowdGrader lets students submit and 
collaboratively grade solutions to homework assignments.
The tool permitted both ranking and grading of homework.
However, de Alfaro and Shavlovsky found that
students much preferred to grade than
to rank. They expressed uneasiness in 
ranking their peers, perceiving ranking as a 
blunt tool compared to grading. 
At the heart of CrowdGrader is the {\em Vancouver} algorithm
for combining peer grades. 
\myOmit{
In each iteration, this 
algorithm computes a consensus estimate of the 
grade of each submission, weighing grades 
according to the (estimated) accuracy of each student; 
the consensus estimates are then used to update the estimated 
accuracy of the students. The update attempts to
minimize variance. 

On synthetic data, 
the Vancouver algorithm perform much better
than simple baselines like the average or median. 
However on some real-world coursework, the results were 
more mixed. 
To encourage good reviewing, the final mark of 
each student was
computed as a weighted average of the consensus grade
and a reviewing grade  which measures the effort and
accuracy of their reviews. Typically this gave
three times the weight to the consensus grade
than to the reviewing grade. 
}
There are two significant differences between
the Vancouver algorithm and the PeerRank rule.
First, the Vancouver algorithm measures variance
in grades whilst PeerRank (like the NSF rule)
measures the absolute deviation. A rule based on variance 
will tend to penalise inexperienced agents more greatly.
Second, the reward term in the Vancouver algorithm 
is added after a fixed point is reached, whilst in the PeerRank rule,
it is part of the fixed point calculation. 
We conjecture that it is more robust
to include it in the fixed point calculation
when there is significant variation in the accuracy
of grades assigned by a single agent. 


There are a number of closely related problems
to ours in the social choice literature. For example,
Holzman and Moulin have
studied a related problem
in which a set of agents wish 
to select one amongst them to receive a 
prize \cite{hmeconometrica2013}. 
A fundamental assumption of this work is that
nominations are impartial: your message never 
influences whether you win the prize.
In our setting, such an
assumption is less appropriate. We 
want your evaluation of the work of another agent
to influence your evaluation. 
There are several reasons behind this
change. First, your ability to evaluate the
work of other agents measures in part your 
command of the subject being examined. 
Second, you will be incentivized to
grade accurately by a better final grade.
If your message cannot influence your evaluation,
then you have no incentive to provide good evaluations. 
For this reason, even if we extend
the sort of methods proposed by Holzman and Moulin to the
task of ranking, 
the
starting assumptions are very different. 

Another related problem is ``selection from the selectors''
\cite{selectingselectors}. The goal here is to select a subset
of $k$ agents from a group (e.g. to select a subcommittee).
The problem of awarding a prize 
from a group of peers can
be seen as the special case of $k=1$. 
As approval voting is not impartial,
Alon, Fischer, Procaccia and Tennenholtz 
look for impartial rules that approximate 
approval voting (that is, guarantee that
the total approval scores of the $k$ winners
are within a fixed fraction of the optimal
answer). Again, a difference with this 
work is that we are not trying to achieve
or approximate impartiality.

A closely related problem is the division of
cash between a group of partners \cite{dividedollar}.
Each partner cares selfishly about their
share but is supposed to be disinterested
about the distribution of the money that
he or she does not get. Partners rate the relative
contributions of the other partners. With
four or more partners, there exist symmetric
and impartial division rules. By comparison,
whilst our PeerRank rule is symmetric, it
is not designed to be impartial. The grades assigned
by an agent
can definitely influence their final grade. 

\myOmit{

Finally, an interesting question is how our work relates
to the sizeable literature on the (strategyproof) 
allocation of goods. Some of the properties
that we prove of the PeerRank rule (like
Symmetry, No Discrimination and No Dummy) are 
also discussed in this literature (e.g. \cite{nodummy1,nodummy2}).
On the one hand, we are not directly interested
in strategyproofness. We want
an agent to be able to improve their
outcome (by rating themselves and their peers
more accurately). On the other hand, 
we want their ability to manipulate
the outcome to provide only an incentive
to report sincere grades. 
We expect that there may be important
impossibility theorems to be found about
peer assessment similar to those in these
related literatures. 
}

\section{Conclusions}

We have proposed the PeerRank method for peer assessment.
The PeerRank method weights 
grades by the grades of the grading agents.
In addition, it rewards agents for 
grading well and penalises those that
grade poorly. As the grades of an agent
depend on the grades of the grading agents,
and as these grades themselves depend on the
grades of other agents, the PeerRank method is defined
by a fixed point equation similar
to the PageRank method. 
We have identified some formal properties of the PeerRank method,
discussed some examples, and evaluated the performance
on some synthetic data. 
The method reduces the error in grade predictions by a factor of 2 
or more in many cases over the natural baseline of 
simply averaging peer grades. 
As the method
favours consensus, it is most suited
to domains where there are objective
answers but the number of agents is too large
for anything but peer grading. 

There are many possible extensions. For example,
we might consider peer assessments where agents
only grade a subset of each other. The PeerRank
rule lifts easily to this case. 
As a second example, we
might permit external calibration by having some
agents graded externally. As a third example, we 
might consider peer assessment when agents
order rather than grade. They might
be willing to say ``agent $a$ should
be graded higher than agent $b$'', or 
``agent $a$ should receive a similar grade as agent $b$'' 
but the grading agents might be 
less willing to give an absolute grade without seeing 
the work of all other agents.
Another interesting direction would
be to return a distribution or interval of 
grades, reflecting the uncertainty in the 
estimate. This could be calculated based on
the intermediate grades seen before the fixed
point is reached.

\bibliographystyle{ecai2014}
\bibliography{/Users/twalsh/Documents/biblio/a-z,/Users/twalsh/Documents/biblio/a-z2,/Users/twalsh/Documents/biblio/pub,/Users/twalsh/Documents/biblio/pub2}

\begin{thebibliography}{10}

\bibitem{selectingselectors}
N.~Alon, F.~Fischer, A.~Procaccia, and M.~Tennenholtz, `Sum of us:
  Strategyproof selection from the selectors', in {\em Proceedings of the 13th
  Conferene on Theoretical Aspects of Rationality and Knowledge}, pp. 101--110,
  (2011).

\bibitem{crowdgrader}
L.~de~Alfaro and M.~Shavlovsky, `Crowdgrader: Crowdsourcing the evaluation of
  homework assignments', Technical Report 1308.5273, arXiv.org, (August 2013).

\bibitem{dividedollar}
G.~DeClippel, H.~Moulin, and N.~Tideman, `Impartial division of a dollar', {\em
  Journal of Economic Theory}, {\bf 139},  176--191, (2007).

\bibitem{hmeconometrica2013}
R.~Holzman and H.~Moulin, `Impartial nominations for a prize', {\em
  Econometrica}, {\bf 81}(1),  173--196, (2013).

\bibitem{kpdjeb95}
P.M. Kerr, K.H. Park, and B.R. Domazlicky, `Peer grading of essays in a
  principles of microeconomics course', {\em Journal of Education for
  Business}, {\bf 70}(6),  357--361, (1995).

\bibitem{nsf1}
M.R. Merrifield and D.G. Saari, `Telescope time without tears: a distributed
  approach to peer review', {\em Astronomy \& Geophysics}, {\bf 50}(4),
  4.16--4.20, (2009).

\bibitem{nsf2}
P.~Naghizadeh and M.~Liu, `Incentives, quality, and risks: A look into the
  {NSF} proposal review pilot', Technical Report 1307.6528, arXiv.org, (July
  2013).

\bibitem{pagerank}
L.~Page, S.~Brin, R.~Motwani, and T.~Winograd, `The {PageRank} citation
  ranking: Bringing order to the web.', Technical Report 1999-66, Stanford
  InfoLab, (November 1999).

\bibitem{edm2013}
Chris Piech, Jonathan Huang, Zhenghao Chen, Chuong Do, Andrew Ng, and Daphne
  Koller, `Tuned models of peer assessment in {MOOCs}', in {\em Proceedings of
  The 6th International Conference on Educational Data Mining (EDM 2013)},
  (2013).

\bibitem{peas}
J.~Singh, K.~Jain, N.~Vedula, P.~Mathur, S.~Agrawal, and P.~Agraewal.
\newblock {PEAS}: Peer expect autograde self: Peer evaluation system for
  {MOOC}, 2013.
\newblock Poster, Fundamental Research Group, Dept of CSE, IIT Bombay.

\end{thebibliography}

\end{document}